\PassOptionsToPackage{section}{placeins}
\documentclass[]{bytedance}
\input{glyphtounicode}
\usepackage[toc,page,header]{appendix}
\usepackage{minitoc}
\usepackage{amsfonts}
\usepackage{amssymb}
\usepackage{tabularx}
\usepackage{listings}
\usepackage{xcolor}
\usepackage{cancel}

\usepackage{tabulary,multirow,xspace}
\usepackage{fixmath,mathtools,nicefrac,mmstyle}
\usepackage{subcaption}
\usepackage{caption}
\usepackage[misc]{ifsym} 
\usepackage{colortbl}

\usepackage{wrapfig}
\usepackage{multicol}
\usepackage[most]{tcolorbox}
\usepackage{pifont}
\usepackage{colortbl, booktabs, multirow, array}

\definecolor{codegreen}{rgb}{0,0.6,0}
\definecolor{codegray}{rgb}{0.5,0.5,0.5}
\definecolor{codepurple}{rgb}{0.58,0,0.82}
\definecolor{backcolour}{rgb}{0.95,0.95,0.92}
\definecolor{boxblue}{RGB}{57,89,163}
\definecolor{boxbluebg}{RGB}{230,237,250} 

\lstdefinestyle{mystyle}{
    backgroundcolor=\color{backcolour},   
    commentstyle=\color{codegreen},
    keywordstyle=\color{magenta},
    numberstyle=\tiny\color{codegray},
    stringstyle=\color{codepurple},
    basicstyle=\ttfamily\footnotesize,
    breakatwhitespace=false,         
    breaklines=true,                 
    captionpos=b,                    
    keepspaces=true,                 
    numbers=none,                    
    numbersep=5pt,                  
    showspaces=false,                
    showstringspaces=false,
    showtabs=false,                  
    tabsize=2
}
\definecolor{mygray1}{gray}{.95}
\definecolor{mygray2}{gray}{.9}
\definecolor{mygray3}{gray}{.95}
\usepackage{pifont}

\newlength\savewidth
\newcolumntype{x}[1]{>{\centering\arraybackslash}p{#1pt}}

\newcommand{\app}{\raise.17ex\hbox{$\scriptstyle\sim$}}

\usepackage{xcolor}
\usepackage{graphicx}
\usepackage{amssymb}
\usepackage{pifont}
\usepackage{floatrow}
\usepackage{amsmath} 
\usepackage{float}
\usepackage{wrapfig}
\usepackage{multirow}
\usepackage{tcolorbox}
\tcbuselibrary{breakable, skins, raster}
\usepackage{listings}
\usepackage{listings}

\usepackage{algorithm}
\usepackage{algorithmic}
\definecolor{myblue}{RGB}{210, 225, 255}
\definecolor{mytextblue}{RGB}{51, 161, 201}
\definecolor{mypurple}{RGB}{218, 112, 214}

\definecolor{commentgreen}{rgb}{0.1, 0.4, 0.1}
\definecolor{keywordblue}{rgb}{0.1, 0.1, 0.7}
\definecolor{stringred}{rgb}{0.7, 0.1, 0.1}

\lstdefinestyle{mystyle}{
    commentstyle=\color{commentgreen},
    keywordstyle=\color{keywordblue},   
    stringstyle=\color{stringred},
    basicstyle=\ttfamily\scriptsize, 
    breaklines=true,
    keepspaces=true,
    showstringspaces=false,
    frame=none,                     
    language=Python, 
}

\newcommand{\name}{SemComp-Bench}
\title{\name{}: Benchmarking Semantic Task Completion in Video Generation}

\author{
Keyu Tu$^1$\quad
Zhuowei Chen$^1$\quad
Mengqi Huang$^{1,2,\dagger}$\quad
\\
Yuxin Wang$^2$\quad
Jiahao Zhu$^3$\quad
Zhendong Mao$^1$\quad
Yongdong Zhang$^1$
}
\affiliation{
$^1$University of Science and Technology of China\quad
$^2$FrameX.AI\quad
$^3$Sun Yat-sen University \\[2pt]
{\normalfont\small $^\dagger$Corresponding author}
}

\abstract{
We introduce \textbf{Semantic Task Completion Video Generation}, an outcome-oriented video generation task.
Under this formulation, success requires both achievement of the intended outcome and semantic grounding.
Semantic grounding characterizes the correspondence between the reference image and the generated outcome in terms of high-level semantics relevant to the task.
Evaluation focuses on the generated outcome and requires neither the presentation of a complete sequence of intermediate task steps nor conventional appearance consistency with the reference image.
To support systematic evaluation, we construct \textbf{SemComp-Data}, an evaluation dataset covering six domains.
Each instance comprises a reference image, a detailed instruction, a brief instruction, and an outcome-centric video clip.
A scalable four-stage curation pipeline converts raw videos into standardized SemComp-Data instances.
We further introduce \textbf{SemComp-Bench}, an evaluation protocol that uses a vision-language model (VLM) to answer structured binary questions.
SemComp-Bench reports the \textbf{OA Score} and the \textbf{GR Score} for Outcome Achievement and Generation Reliability, respectively.
Experiments on representative video generation models show that achieving intended outcomes while maintaining task-relevant semantic grounding in reference images remains challenging.
}

\date{\today}
\checkdata[Project Page]{\url{https://SemComp-Bench.github.io}}
\correspondence{Mengqi Huang at \email{huangmq@ustc.edu.cn}}

\begin{document}
\maketitle

\section{Introduction}

Recent video generation models~\cite{wan2025wan,seedance2025,hunyuanvideo2025} achieve impressive visual fidelity and temporal coherence.
However, their ability to achieve an instructed outcome while maintaining \textbf{high-level semantic grounding} remains underexplored.
We formulate this problem as \textbf{Semantic Task Completion Video Generation}.
Here, grounding preserves task-relevant semantic relationships and reference attributes while allowing unrelated attributes to change.
\begin{figure}[ht]
    \centering
    \includegraphics[width=1.0\linewidth]{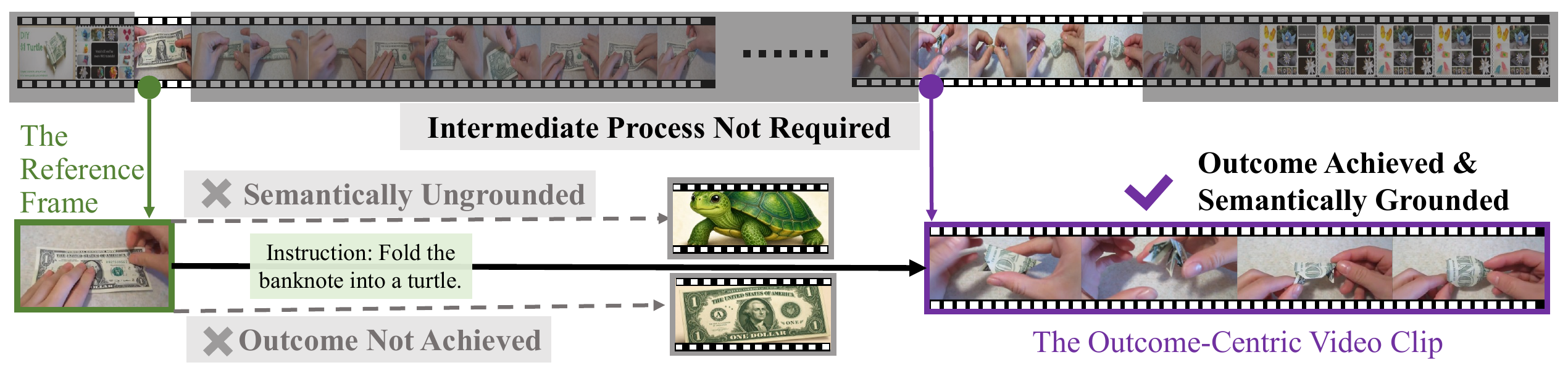}
    \caption{Demonstration of Semantic Task Completion Video Generation, which focuses not on the transformation process but solely on outcome achievement and semantic grounding.}
    \label{fig:Demo1}
\end{figure}
For example, given a banknote image and the instruction ``Fold the banknote into a turtle,'' a successful video must visibly present the referenced banknote as turtle-shaped origami, as illustrated in Fig.~\ref{fig:Demo1}.
Intermediate folding steps are unnecessary, but the result must remain grounded in the input banknote rather than replacing it with an unrelated turtle.

\begin{figure*}[h]
    \centering
    \includegraphics[width=1.0\linewidth]{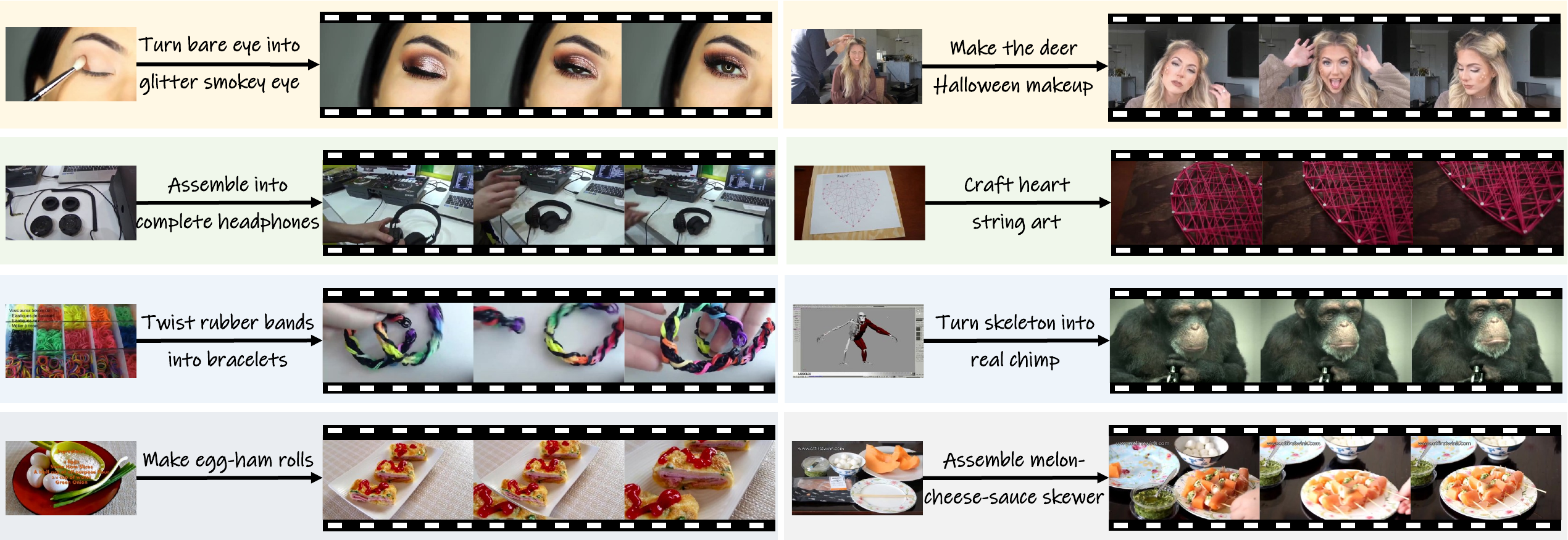}
    \caption{Representative SemComp-Data instances. Each example shows a reference frame, its outcome-centric video clip, and the brief instruction for compact presentation; every instance also includes a detailed instruction.}
    \label{fig:Demos}
\end{figure*}

Existing datasets and benchmarks emphasize visual fidelity, temporal coherence, and subject consistency, but rarely assess outcome achievement jointly with high-level semantic grounding.
We therefore construct \textbf{SemComp-Data}, an evaluation dataset of image-text-video triplets curated from \textbf{full-context} real-world videos. 
Such full-context curation naturally ensures task feasibility while enabling fine-grained visual alignment.
Each instance pairs the same reference and outcome with brief and detailed instructions describing the intended result at different levels of specificity.
The curation pipeline comprises \textbf{Candidate Filtering}, \textbf{State Mining}, \textbf{Video Extension}, and \textbf{Instruction Structuring}.
Candidate Filtering screens raw videos, State Mining localizes and verifies reference--outcome frames, and Video Extension extracts an outcome-centric clip around the localized outcome. Instruction Structuring then identifies alignment constraints and produces instructions.
Figure~\ref{fig:Demos} presents representative SemComp-Data instances.

Building on SemComp-Data, \textbf{SemComp-Bench} uses evidence-grounded binary VLM judgments to assess two complementary dimensions: \textbf{Outcome Achievement} and \textbf{Generation Reliability}, reported as the OA Score and GR Score.
The evaluator answers predefined questions and supports each answer with visual evidence, yielding focused and interpretable judgments.
OA measures outcome realization and semantic grounding, with safeguards for task-relevant entity consistency and global visual continuity.
GR complements it by assessing physical violations, blur, rendering artifacts, local instability, and corrupted text or interface elements.
Together, the scores support systematic evaluation and criterion-level failure diagnosis.
The main contributions of this work are summarized as follows:
\begin{itemize}
    \item We formulate \textbf{Semantic Task Completion Video Generation}, a new task that requires generated videos to achieve instructed outcomes while preserving task-relevant semantic relationships with reference images.

    \item We construct \textbf{SemComp-Data} through a scalable, low-cost pipeline that converts raw videos into image-text-video triplets with paired brief and detailed instructions.

    \item We introduce \textbf{SemComp-Bench}, a VLM-based evaluation protocol that measures Outcome Achievement and Generation Reliability using interpretable, evidence-grounded binary judgments.
\end{itemize}

\section{Related Work}

\subsection{Video Generation Benchmarks}
Video generation benchmarks evaluate visual fidelity, temporal coherence, and prompt adherence~\cite{liu2024evalcrafter,feng2024tc,huang2024vbench,huang2025vbench++,zheng2025vbench}. Personalized and controllable benchmarks further assess subject identity and intended motion~\cite{yuan2025opens2v,ling2025vmbench}, while recent work examines physical plausibility, causal consistency, and commonsense constraints~\cite{li2026worldmodelbench,liu2026rise,bansal2025videophy,yuan2025likephys,zhao2026worldolympiad,bai2025impossible}.
Together, these benchmarks cover generation quality, identity, motion, and physical plausibility. However, they do not directly test whether a video achieves an outcome jointly defined by an instruction and reference context while preserving task-relevant semantic grounding, motivating SemComp-Bench.

\subsection{Video Generation Models}
Current models increasingly support high-quality, instruction-following generation.
Open-source systems such as Wan2.2~\cite{wan2025wan}, CogVideoX~\cite{yang2025cogvideox}, HunyuanVideo~\cite{kong2024hunyuanvideo}, and Pyramid Flow~\cite{jin2025pyramidal} support text- and image-conditioned generation, while LTX~\cite{hacohen2024ltx,hacohen2026ltx} emphasizes efficiency.
Closed-source systems, including Sora, Veo, MovieGen, Runway, Kling, Seedance, Hailuo, and Pika, further support multimodal conditioning, multi-shot narratives, camera control, and editing~\cite{sora2024,veo2025,moviegen2024,runway2025,kling2025,seedance2025,hailuo2025,pika2025}.
Together, these models show strong synthesis and control, but their outcome achievement under task-relevant reference grounding remains insufficiently understood.

\section{SemComp-Data Construction}\label{sec:pipeline}
    \begin{figure*}[!t]
    \centering
    \includegraphics[width=0.95\textwidth]{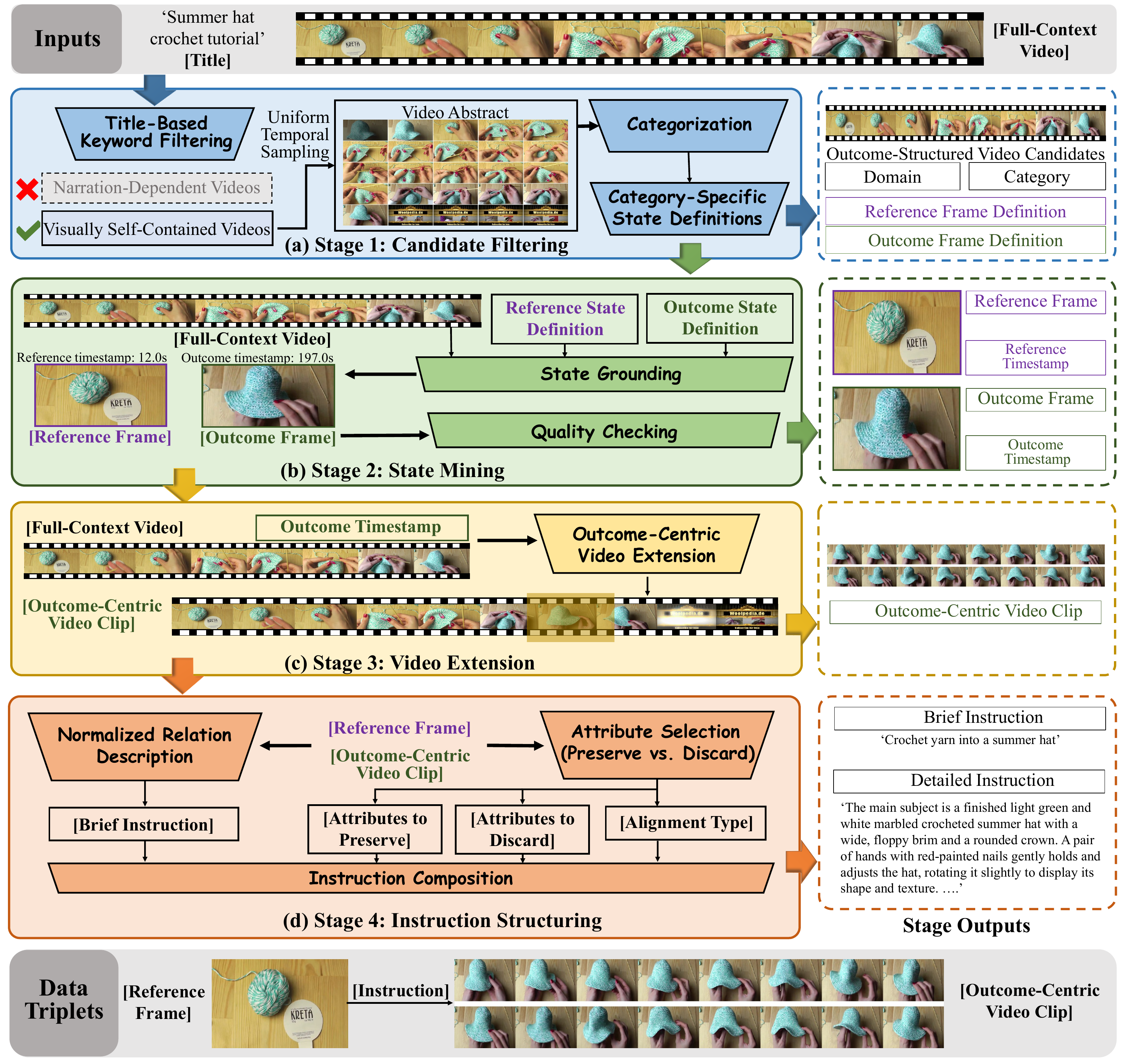}
    \caption{Overview of the four-stage SemComp-Data curation pipeline.
    Panels (a)--(d) show Candidate Filtering, State Mining, Video Extension,
    and Instruction Structuring, respectively. The right column shows the
    corresponding stage outputs, and the bottom row presents the final
    data triplet.}
    \label{fig:framework}
    \end{figure*}

Starting from \textbf{full-context} videos in Koala-36M~\cite{wang2025koala}, we construct SemComp-Data as a collection of structured image-text-video triplets $x_i=(r_i,\mathcal{I}_i,o_i)$, where $\mathcal{I}_i=(i_i^{\mathrm{brief}},i_i^{\mathrm{detailed}})$ is an aligned instruction pair describing the same intended outcome at two levels of specificity.
The reference frame $r_i$ defines the initial context, the paired instructions specify the desired outcome, and the outcome-centric video clip $o_i$ depicts successful completion.
Both $r_i$ and $o_i$ are drawn from the same task instance, ensuring that the visual elements are genuinely associated rather than independently collected.
This full-context construction provides evidence that the intended outcome is realizable and visually verifiable for the given reference.
As illustrated in Fig.~\ref{fig:framework}, we develop \textbf{a four-stage curation pipeline} to construct standardized evaluation instances from full-context videos.
The pipeline supports scalable data construction without requiring the manual design of individual tasks or the production of corresponding outcome-centric videos.
The following subsections describe each stage in detail.

\subsection{Stage 1: Candidate Filtering}\label{sec:candidate filtering}
We use the Koala-36M~\cite{wang2025koala} dataset as the source pool, owing to its broad thematic coverage and strong representation of real-world scenarios.
Because SemComp-Data relies on visually observable outcomes, we first apply \textbf{Title-Based Keyword Filtering} to exclude narration-dependent videos whose titles contain lexical patterns such as \textit{talk show}, \textit{interview}, and \textit{news}, thereby retaining candidates that are more likely to be visually self-contained, as shown in Fig.~\ref{fig:framework}(a).
We uniformly sample frames from each retained video and arrange them into a mosaic representation, termed the \textit{video abstract}, which provides a compact visual summary for efficient categorization.
Using these abstracts, an advanced VLM assigns each video to one of six domains and an associated task category; videos lacking sufficient visual evidence for reliable categorization are labeled as \emph{Uncertain} and discarded.
The retained domains are \emph{Arts and Precision}, \emph{Beauty and Fashion}, \emph{Crafts and DIY}, \emph{Food and Cooking}, \emph{Gardening and Pets}, and \emph{Sports and Fitness}.
For each task category, we manually define the reference and outcome states according to its characteristic completion pattern. 
These definitions specify what constitutes the initial context and the completed outcome for downstream state mining.
The complete state definitions and the keyword list used for title-based filtering are provided in the supplementary material.

\subsection{Stage 2: State Mining}\label{sec:state mining}
As shown in Fig.~\ref{fig:framework}(b), this stage uses the manually specified category-specific state definitions from Stage 1 to mine reliable reference--outcome frame pairs through two steps: \textbf{State Grounding} and \textbf{Quality Checking}.
We formulate State Grounding as frame-level timestamp localization rather than direct outcome-segment localization, allowing the VLM to focus on representative visual evidence of the reference and outcome states without modeling the intermediate process or resolving ambiguous segment boundaries. 
Given a full-context video and its state definitions, the VLM first identifies the demonstrated task and expected outcome and then locates the timestamps that best match the two state definitions.
The reference and outcome frames at these timestamps are extracted for subsequent processing.
The extracted frame pair is then verified through QA-based Quality Checking, which serves as a conservative consistency screen.
The VLM is provided with the two unlabeled frames and the task description obtained during State Grounding and is asked to identify which frame represents the outcome. 
A valid pair should allow the outcome frame to be identified unambiguously.
If the predicted outcome does not match the grounded outcome frame, the pair is conservatively discarded, as the disagreement may reflect an ambiguous reference--outcome relationship, insufficient visual evidence, or VLM error.
The remaining reference and outcome frames, together with their timestamps, constitute the outputs of this stage.

\subsection{Stage 3: Video Extension}\label{sec:video extension}
As shown in Fig.~\ref{fig:framework}(c), this stage takes the full-context video and the verified outcome timestamp from Stage 2 as inputs.
Following the shot detection and same-scene merging procedure of Panda-70M~\cite{chen2024panda}, we partition the full-context video into segments with consistent camera viewpoints.
Using the verified timestamp as a temporal anchor, we extract an outcome-centric clip ($o_i$) by selecting the core segment containing the outcome frame and merging visually consistent neighboring segments until a minimum duration of \(3\,\mathrm{s}\) is reached.
Compared with directly prompting a VLM to localize the outcome interval, this timestamp-anchored strategy more effectively centers the extracted clip on the completed outcome while offering greater efficiency and lower cost.
The clips have an average duration of approximately \(4.03\,\mathrm{s}\) in SemComp-Data, providing sufficient temporal evidence of the completed outcome while minimizing intermediate processes and unrelated content.
The final outcome-centric clip constitutes the output of this stage.

\subsection{Stage 4: Instruction Structuring}\label{sec:instruction structuring}
As shown in Fig.~\ref{fig:framework}(d), this stage takes the reference frame and outcome-centric video clip as inputs and produces two aligned instruction variants---a brief instruction $i_i^{\mathrm{brief}}$ and a detailed instruction $i_i^{\mathrm{detailed}}$---through \textbf{Normalized Relation Description}, \textbf{Attribute Selection}, and \textbf{Instruction Composition}.
In preliminary experiments, directly prompting the VLM with the two inputs often produces overly verbose instructions containing incidental details, such as brand names and on-screen text. 
To capture the essential reference--outcome relation, Normalized Relation Description constrains the VLM to generate a brief instruction $i_i^{\mathrm{brief}}$ of no more than 30 words following the template [Verb] + [Main Subject in $r_i$] + [Preposition] + [Main State in $o_i$].

In parallel, Attribute Selection determines which reference attributes should be preserved and which can be discarded during outcome-centric video generation. 
Because the task-relevant attributes vary across categories, the VLM first selects an applicable alignment type from four options: \emph{Object Element}, \emph{Person Identity}, \emph{Object Appearance}, and \emph{Scene}. 
For example, person identity and task-relevant appearance cues should be preserved when the reference and outcome depict the same person with different makeup or hairstyles, whereas the spatial layout and design structure should be retained when transforming a blueprint into its corresponding physical result.
Based on the selected alignment type, the VLM identifies the attributes to preserve and discard. 
The definitions and available options for the alignment types are provided in the supplementary material.

Instruction Composition expands the brief instruction $i_i^{\mathrm{brief}}$ into the detailed instruction $i_i^{\mathrm{detailed}}$ by combining (i) the selected alignment type and preserve/discard attributes, which specify the reference-grounding constraints, and (ii) fine-grained visual characteristics that describe the completed outcome, including its background, lighting, colors, shapes, and component-level details.
These outcome characteristics do not constitute reference-appearance constraints unless they are explicitly selected for preservation.
The resulting detailed instruction jointly specifies the desired outcome and its reference-grounding requirements, making it suitable for evaluation and potentially useful for future task-specific training.
The two variants describe the same task instance at different levels of specificity, enabling controlled evaluation of instruction specificity without changing the underlying reference or target outcome. The brief instruction better reflects typical user prompting habits and therefore provides a more challenging evaluation setting.

\begin{figure*}[t]
    \centering
    \includegraphics[width=1.0\textwidth]{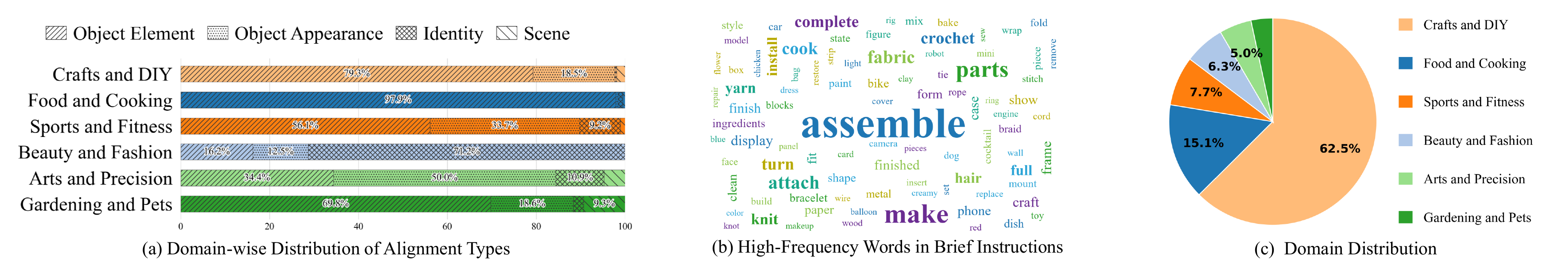}
    \caption{Statistics of SemComp-Data: (a) distribution of alignment types across domains, with distinct textures denoting different types; (b) high-frequency words in the instructions; and (c) distribution of evaluation instances across the six domains.}
\label{fig:data_summary}
\end{figure*}

Figure~\ref{fig:data_summary} summarizes the alignment-type and domain distributions of SemComp-Data, together with high-frequency words in the brief instructions.

\begin{table*}[!t]
    \centering
    \small
    \setlength{\tabcolsep}{4pt}
    {
    \begin{tabular}{
        >{\raggedright\arraybackslash}m{0.38\linewidth}
        *{4}{>{\centering\arraybackslash}m{0.095\linewidth}}
        >{\centering\arraybackslash}m{0.095\linewidth}
    }
    \toprule
    Models
    & $A_{\mathrm{or}}$
    & $A_{\mathrm{sg}}$
    & $A_{\mathrm{gec}}$
    & $A_{\mathrm{gvc}}$
    & \textbf{OA Score} \\
    \midrule

    Seedance 2.0~\cite{bytedance2026seedance20}
    & \underline{0.839}
    & \textbf{0.744}
    & 0.444
    & 0.594
    & 20.0\% \\

    Wan2.2-TI2V-5B~\cite{wan2025wan}
    & 0.589
    & 0.400
    & \textbf{0.689}
    & \textbf{0.922}
    & 23.3\% \\

    Wan2.2-I2V-A14B~\cite{wan2025wan}
    & 0.800
    & 0.528
    & \underline{0.628}
    & 0.789
    & \underline{28.3\%} \\

    CogVideoX1.5-5B-I2V~\cite{yang2025cogvideox}
    & 0.550
    & 0.389
    & 0.506
    & 0.744
    & 14.4\% \\

    SkyReels-V2-I2V-14B-720P~\cite{chen2025skyreelsv2}
    & 0.733
    & 0.489
    & 0.522
    & 0.772
    & 22.8\% \\

    HY$^\dagger$-1.5-720P-I2V
    & \textbf{0.878}
    & \underline{0.706}
    & 0.583
    & \underline{0.794}
    & \textbf{37.8\%} \\

    Phantom-1.3B~\cite{liu2025phantom}
    & 0.539
    & 0.356
    & 0.322
    & 0.511
    & 3.9\% \\

    \bottomrule
    \end{tabular}
    }
    \caption{SemComp-Core Outcome Achievement with detailed instructions. $A_{\mathrm{or}}$, $A_{\mathrm{sg}}$, $A_{\mathrm{gec}}$, and $A_{\mathrm{gvc}}$ denote pass rates for outcome realization, semantic grounding, grounded entity consistency, and global visual continuity, respectively. The OA Score is the joint pass rate across four criteria. $^\dagger$HY denotes HunyuanVideo~\cite{hunyuanvideo2025}. Bold and underlined values indicate the best and second-best values, respectively.}
    \label{tab:oa_results}
\end{table*}

\section{SemComp-Bench Evaluation}\label{sec:semcomp bench}

Building on SemComp-Data, SemComp-Bench evaluates generated videos through structured binary questions, each targeting a specific criterion for interpretable assessment and failure diagnosis.
The criteria are organized into two complementary dimensions: \textbf{Outcome Achievement (OA)}, which measures outcome realization, reference grounding, and task-relevant continuity; and \textbf{Generation Reliability (GR)}, which assesses visual, physical, and temporal reliability.
Their aggregate results are reported as the \textbf{OA Score} and \textbf{GR Score}, respectively.
For a single evaluation run on a set of $N$ samples, let $a_c^i, g_c^i\in\{0,1\}$ denote the binary criterion scores of sample $i$ for criterion $c$ in the Outcome Achievement and Generation Reliability dimensions, respectively, where $1$ indicates a pass and $0$ otherwise.
The full set of binary questions is provided in the supplementary material.

\subsection{Outcome Achievement}

For each sample $i$, outcome achievement is evaluated using four criteria: outcome realization, semantic grounding, grounded entity consistency, and global visual continuity.
Their binary criterion scores are $a_{\mathrm{or}}^i$, $a_{\mathrm{sg}}^i$, $a_{\mathrm{gec}}^i$, and $a_{\mathrm{gvc}}^i$, respectively.
All four criteria are evaluated on the same temporally ordered sequence of 27 frames uniformly sampled from each generated video.
Outcome realization and semantic grounding additionally use the reference image and instruction, whereas grounded entity consistency and global visual continuity use only the sampled sequence.
The corresponding dataset-level pass rates are $A_{\mathrm{or}}$, $A_{\mathrm{sg}}$, $A_{\mathrm{gec}}$, and $A_{\mathrm{gvc}}$, where
$A_c=\frac{1}{N}\sum_{i=1}^{N}a_c^i$.

The outcome realization criterion assesses whether the generated video visibly reaches the completed state specified by the instruction at a coarse semantic level, independent of fine-grained correspondence to the reference image.
The semantic grounding criterion assesses whether the realized outcome preserves or modifies task-relevant entities and attributes according to the reference-instruction pair.
The grounded entity consistency criterion assesses whether task-relevant key entities remain semantically identifiable throughout the sampled sequence, without unexplained disappearance, replacement, or unintended drift in identity, material, appearance, or task-relevant structure.
The global visual continuity criterion examines the temporally ordered sampled sequence for abrupt global switches in scene, viewpoint, layout, background, or composition.
A common, but not exclusive, failure case occurs when a reference-matching initial frame is followed by a visually disconnected continuation.
These two criteria serve as validity safeguards for the generated outcome and do not assess the completeness or procedural correctness of intermediate task steps.

All four indicators are pass-coded, with $1$ denoting that the corresponding criterion is satisfied. For the two failure-oriented questions on entity inconsistency and global visual discontinuity, a ``No'' response is mapped to $1$.
Overall success is conjunctive: the sample-level Outcome Achievement indicator $A^i$ equals $1$ only when sample $i$ passes all four criteria.
The dataset-level \textbf{OA Score} is the mean of $A^i$ across all samples:
\begin{equation}
\begin{gathered}
A^i
=
a_{\mathrm{or}}^i
a_{\mathrm{sg}}^i
a_{\mathrm{gec}}^i
a_{\mathrm{gvc}}^i
\in\{0,1\},
\\
\text{OA Score}
=
\frac{1}{N}\sum_{i=1}^{N}A^i.
\end{gathered}
\label{eq:oa_score}
\end{equation}

\begin{table*}[t]
    \centering
    \small
    \setlength{\tabcolsep}{3pt}
    {
    \begin{tabular}{
        >{\raggedright\arraybackslash}m{0.38\linewidth}
        *{5}{>{\centering\arraybackslash}m{0.075\linewidth}}
        >{\centering\arraybackslash}m{0.09\linewidth}
    }
    \toprule
    Models
    & $G_{\mathrm{pp}}$
    & $G_{\mathrm{vc}}$
    & $G_{\mathrm{afr}}$
    & $G_{\mathrm{wsc}}$
    & $G_{\mathrm{ti}}$
    & \textbf{GR Score} \\
    \midrule

    Seedance 2.0~\cite{bytedance2026seedance20}
    & 0.883
    & \textbf{0.994}
    & \textbf{0.994}
    & \textbf{0.739}
    & \textbf{0.978}
    & \textbf{91.8\%} \\

    Wan2.2-TI2V-5B~\cite{wan2025wan}
    & 0.794
    & \underline{0.978}
    & 0.889
    & \underline{0.722}
    & 0.889
    & 85.4\% \\

    Wan2.2-I2V-A14B~\cite{wan2025wan}
    & \underline{0.911}
    & 0.972
    & \underline{0.967}
    & 0.672
    & \underline{0.928}
    & \underline{89.0\%} \\

    CogVideoX1.5-5B-I2V~\cite{yang2025cogvideox}
    & \textbf{0.944}
    & 0.811
    & 0.772
    & 0.583
    & 0.828
    & 78.8\% \\

    SkyReels-V2-I2V-14B-720P~\cite{chen2025skyreelsv2}
    & 0.778
    & 0.922
    & 0.739
    & 0.328
    & 0.789
    & 71.1\% \\

    HY$^\dagger$-1.5-720P-I2V
    & 0.800
    & 0.939
    & 0.883
    & 0.472
    & 0.861
    & 79.1\% \\

    Phantom-1.3B~\cite{liu2025phantom}
    & 0.778
    & 0.972
    & 0.856
    & 0.506
    & 0.728
    & 76.8\% \\
    
    \bottomrule
    \end{tabular}
    }
    \caption{SemComp-Core Generation Reliability with detailed instructions. Each $G_c$ denotes the pass rate for a reliability criterion, and the GR Score is the mean across five criteria. $^\dagger$HY denotes HunyuanVideo~\cite{hunyuanvideo2025}. Bold and underlined values indicate the best and second-best values.}
    \label{tab:gr_results}
\end{table*}

\begin{table*}[!t]
    \centering
    \small
    \setlength{\tabcolsep}{3pt}
    {
    \begin{tabular}{
        >{\raggedright\arraybackslash}m{0.25\linewidth}
        >{\centering\arraybackslash}m{0.08\linewidth}
        >{\centering\arraybackslash}m{0.13\linewidth}
        *{4}{>{\centering\arraybackslash}m{0.06\linewidth}}
        >{\centering\arraybackslash}m{0.10\linewidth}
    }
    \toprule
    Model
    & Modality
    & Instruction Type
    & $A_{\mathrm{or}}$
    & $A_{\mathrm{sg}}$
    & $A_{\mathrm{gec}}$
    & $A_{\mathrm{gvc}}$
    & \textbf{OA Score} \\
    \midrule

    \multirow{3}{=}{Wan2.2-A14B~\cite{wan2025wan}}
    & I2V & Detailed
    & 0.800 & 0.528 & 0.628 & 0.789 & 28.3\% \\

    & T2V & Detailed
    & 0.889 & 0.522 & 0.317 & 0.117 & 4.4\% \\

    & T2V & Brief
    & 0.389 & 0.111 & 0.806 & 0.250 & 0.6\% \\
    \midrule

    \multirow{3}{=}{CogVideoX1.5-5B~\cite{yang2025cogvideox}}
    & I2V & Detailed
    & 0.550 & 0.389 & 0.506 & 0.744 & 14.4\% \\

    & T2V & Detailed
    & 0.567 & 0.272 & 0.361 & 0.161 & 5.0\% \\

    & T2V & Brief
    & 0.567 & 0.150 & 0.678 & 0.161 & 0.6\% \\
    \midrule

    \multirow{3}{=}{HY$^\dagger$-1.5-720P}
    & I2V & Detailed
    & 0.878 & 0.706 & 0.583 & 0.794 & 37.8\% \\

    & T2V & Detailed
    & 0.833 & 0.606 & 0.389 & 0.133 & 4.4\% \\

    & T2V & Brief
    & 0.550 & 0.100 & 0.761 & 0.178 & 1.7\% \\
    \bottomrule
    \end{tabular}
    }
    \caption{Outcome Achievement on SemComp-Core across modalities and instruction settings. Within each model family, the I2V and T2V variants use checkpoints of the same parameter scale. Each $A_c$ denotes a criterion-level pass rate, and the OA Score is the joint pass rate. $^\dagger$HY denotes HunyuanVideo~\cite{hunyuanvideo2025}.}
    \label{tab:oa_i2v_t2v_results}
\end{table*}

\subsection{Generation Reliability}

For each sample $i$, the VLM evaluates generation reliability independently of outcome achievement and reference grounding using five binary criteria: physical plausibility, visual clarity, artifact-free rendering, within-scene spatiotemporal coherence, and text and interface integrity.
Their binary criterion scores are $g_{\mathrm{pp}}^i$, $g_{\mathrm{vc}}^i$, $g_{\mathrm{afr}}^i$, $g_{\mathrm{wsc}}^i$, and $g_{\mathrm{ti}}^i$, respectively.
The corresponding dataset-level pass rates are $G_{\mathrm{pp}}$, $G_{\mathrm{vc}}$, $G_{\mathrm{afr}}$, $G_{\mathrm{wsc}}$, and $G_{\mathrm{ti}}$, where $G_c=\frac{1}{N}\sum_{i=1}^{N}g_c^i$.

The physical plausibility criterion assesses visible motions and interactions for clear violations of basic physical principles; it neither requires the instructed transformation process to be shown nor treats an omitted intermediate process as a failure.
The visual clarity criterion identifies content that is difficult to recognize because of blur, improper exposure, insufficient resolution, or unclear scene details.
The artifact-free rendering criterion detects scene-inconsistent synthetic artifacts, such as abnormal noise, corrupted regions, blank patches, or color streaks.
The within-scene spatiotemporal coherence criterion detects low-level rendering instability within continuous scenes, such as flicker, transient duplication, and local geometric deformation, irrespective of task semantics.
Unlike grounded entity consistency, this criterion concerns local rendering stability rather than the semantic persistence of task-relevant entities.
Abrupt global switches across the sampled sequence are evaluated separately by global visual continuity in the Outcome Achievement dimension.
The text and interface integrity criterion detects corrupted or unrecognizable text, numbers, icons, and interface elements.

All five questions are failure-oriented, and the reported indicators are pass-coded: a ``No'' response is mapped to $1$.
The sample-level Generation Reliability score $G^i$ is the arithmetic mean of the five binary criterion scores, and the dataset-level \textbf{GR Score} is the mean of $G^i$ across all samples:
\begin{equation}
\begin{gathered}
G^i
=
\frac{1}{5}
\bigl(
g_{\mathrm{pp}}^i
+g_{\mathrm{vc}}^i
+g_{\mathrm{afr}}^i
+g_{\mathrm{wsc}}^i
+g_{\mathrm{ti}}^i
\bigr)
\in[0,1],
\\
\text{GR Score}
=
\frac{1}{N}\sum_{i=1}^{N}G^i.
\end{gathered}
\label{eq:gr_score}
\end{equation}
Together, the criterion-level pass rates identify specific sources of generation unreliability.

\section{Experiments}
\subsection{Experimental Setup}\label{sec:exp-data}
We randomly sample approximately 20K videos from Koala-36M~\cite{wang2025koala} and process them using the curation pipeline, yielding 1,273 structured evaluation instances.
To construct a domain-balanced evaluation subset, \textbf{SemComp-Core} comprises 60 instances, with 10 selected from each domain via stratified sampling to approximately preserve the within-domain alignment-type distributions.
Each SemComp-Core instance includes both instruction variants.
In analyses of instruction specificity, the paired variants share the same reference image and outcome target, so only the level of instruction specificity varies.
Representative open- and closed-source video generation models~\cite{wan2025wan,yang2025cogvideox,hunyuanvideo2025,bytedance2026seedance20,chen2025skyreelsv2,liu2025phantom} are evaluated on this subset.
All models in Tables~\ref{tab:oa_results} and~\ref{tab:gr_results} use I2V with the reference image and detailed instruction; Wan2.2-TI2V-5B runs in image-conditioned mode.

Open-source models use their official default inference configurations, whereas closed-source models use their exposed default settings; a fixed seed is applied whenever explicit seed control is available.
Under every setting, one 720p video is generated per instance, from which 27 frames are uniformly sampled over the full temporal extent to form a fixed-length, temporally ordered sequence for evaluation.
Every video is scored through three independent VLM calls made at different times, with all criterion-level pass rates and both aggregate scores computed per call using the definitions above; the reported results are arithmetic means of the three corresponding run-level scores.
We instantiate the VLM used in the curation and evaluation experiments as Doubao-Seed-1.8 through the Volcano Engine API~\cite{bytedanceseed2026seed18}, as preliminary experiments indicate that it offers a favorable trade-off between evaluation efficiency and cost.

\subsection{Outcome Achievement Results}\label{sec:OA metric}

Table~\ref{tab:oa_results} reports the OA Score under detailed instructions.
HunyuanVideo-1.5-720P-I2V achieves the highest OA Score of 37.8\%, followed by Wan2.2-I2V-A14B at 28.3\%.
Their results are consistent with relatively balanced pass rates across the four OA criteria, rather than dominance in any single criterion.
Seedance 2.0 performs strongly in outcome realization and semantic grounding, but its lower pass rates on grounded entity consistency and global visual continuity reduce its joint OA Score.
In contrast, Wan2.2-TI2V-5B achieves the highest pass rates on grounded entity consistency and global visual continuity while performing less strongly in outcome realization and semantic grounding.
These complementary performance profiles, together with the fact that the best OA Score remains below 40\%, indicate that robust outcome achievement requires both task fidelity and outcome-video validity.

\FloatBarrier
\subsection{Generation Reliability Results}\label{sec:GR metric}

Table~\ref{tab:gr_results} reports generation reliability under detailed instructions.
Seedance 2.0 achieves the highest GR Score of 91.8\%. Among the evaluated open-source models, Wan2.2-I2V-A14B performs best at 89.0\%.
Both models perform strongly in visual clarity and artifact-free rendering. However, within-scene spatiotemporal coherence remains the primary bottleneck across all models, with pass rates ranging from only 0.328 to 0.739.
This gap suggests that current models can often generate visually convincing individual frames but struggle to maintain stable visual evolution within scenes.
Comparing the GR and OA rankings further indicates that generation reliability does not necessarily correspond to outcome achievement: Seedance 2.0 leads in GR but performs substantially worse in OA, whereas HunyuanVideo-1.5-720P-I2V leads in OA despite a lower GR ranking.
Wan2.2-I2V-A14B ranks second on both metrics, indicating comparatively balanced performance across the two evaluation dimensions.

\FloatBarrier
\subsection{Conditioning Effects}

As shown in Table~\ref{tab:oa_i2v_t2v_results}, I2V variants consistently outperform their T2V counterparts under detailed instructions across all three model families.
The gains mainly arise from improved semantic grounding, grounded entity consistency, and global visual continuity, while outcome-realization pass rates remain broadly comparable and are higher for T2V in two of the three model families.
These findings underscore the indispensable role of reference-image conditioning in semantic task completion, particularly in preserving entity identity, appearance, and structural consistency during complex object interactions and state changes.

Within the T2V setting, detailed instructions consistently achieve higher OA Scores than brief instructions, primarily through better outcome realization and semantic grounding.
In contrast, brief instructions often yield higher grounded entity consistency and global visual continuity, likely because their simpler and less constrained content is easier to generate coherently.
These results reveal a trade-off between instruction specificity and generation difficulty: detailed instructions better define the intended compositional outcome but place greater demands on semantic understanding and coordinated generation, whereas brief instructions are associated with higher outcome-video consistency but lower task-fulfillment rates.

\FloatBarrier
\subsection{Visualization of Generated Videos}

\begin{figure}[!t]
    \centering
    \includegraphics[width=1.0\linewidth]{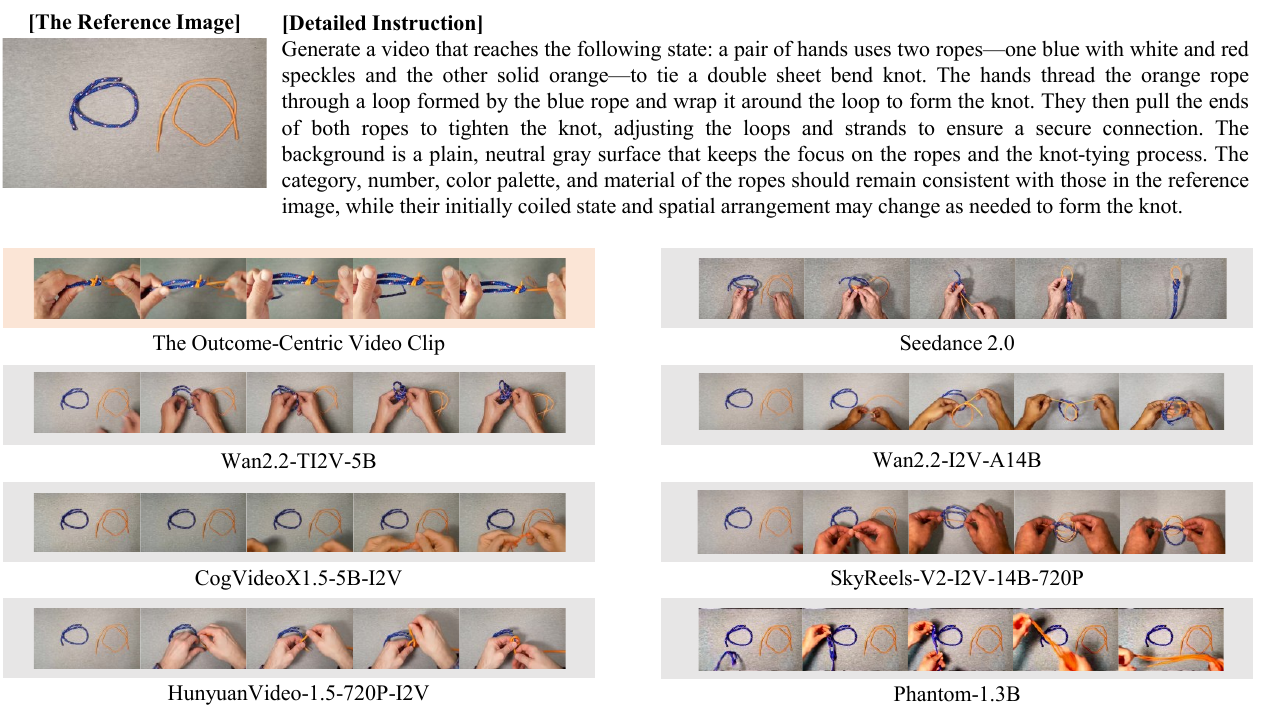}
    \caption{Generated-video examples. The outcome-centric video clip and the reference image are extracted from the same full-context video.}
    \label{fig:modeloutput}
\end{figure}

As shown in Fig.~\ref{fig:modeloutput}, the evaluated models tend to preserve the appearance of the reference image and depict the process of task execution, yet often fail to achieve the intended outcome.
Moreover, generated videos often suffer from reliability issues, including physically implausible transformations and abrupt visual transitions, as illustrated by the failure cases of Wan2.2-I2V-A14B and Phantom-1.3B shown in the figure.
Additional visualizations of generated videos are provided in the supplementary material.

\section{Conclusion}
In this paper, we introduce Semantic Task Completion Video Generation, which requires generated videos to realize an instructed outcome while maintaining semantic grounding in the reference context. 
We develop SemComp-Data through a scalable four-stage curation pipeline, with each instance comprising a reference image, paired brief and detailed instructions, and an outcome-centric video clip.
Crucially, the paired image and clip are extracted from the same full-context video, ensuring task authenticity and achievability as well as fine-grained attribute alignment between the reference and target, while the dataset preserves diverse real-world scenarios from the source collection.
Building on this dataset, SemComp-Bench is the first benchmark to jointly evaluate Outcome Achievement and Generation Reliability.
Experiments with representative video generation models reveal substantial limitations in completing instructed tasks and preserving reference-grounded semantics, indicating that this capability remains largely underexplored.
The potential of SemComp-Data to improve model performance through task-specific training remains to be empirically validated.

\clearpage

\bibliographystyle{plainnat}
\bibliography{main}


\clearpage
\setcounter{section}{0}
\setcounter{figure}{0}
\setcounter{table}{0}
\setcounter{equation}{0}
\setcounter{secnumdepth}{2}
\renewcommand{\thesection}{\Alph{section}}
\renewcommand{\thefigure}{S\arabic{figure}}
\renewcommand{\thetable}{S\arabic{table}}
\renewcommand{\theequation}{S\arabic{equation}}
\renewcommand{\theHsection}{supp.\arabic{section}}
\renewcommand{\theHfigure}{supp.\arabic{figure}}
\renewcommand{\theHtable}{supp.\arabic{table}}
\renewcommand{\theHequation}{supp.\arabic{equation}}
\setcounter{topnumber}{4}
\setcounter{dbltopnumber}{3}
\renewcommand{\topfraction}{0.95}
\renewcommand{\dbltopfraction}{0.95}
\renewcommand{\textfraction}{0.05}

\section*{Supplementary Material}
\phantomsection
\addcontentsline{toc}{section}{Supplementary Material}

The supplementary material provides further details on SemComp-Data and SemComp-Bench.
Specifically, it describes the title-filtering procedure (Sec.~\ref{supp:title-filtering}), the domain and category taxonomy (Sec.~\ref{supp:taxonomy}), category-specific definitions of reference and outcome states (Sec.~\ref{supp:state-definitions}), alignment types and preservation attributes (Sec.~\ref{supp:alignment}), and the complete evaluation prompt (Sec.~\ref{supp:evaluation-prompts}).
It also presents dataset statistics and instances (Sec.~\ref{supp:statistics}), additional generated-video comparisons (Sec.~\ref{supp:qualitative}), and the standard deviations of the reported results (Sec.~\ref{supp:result-std}).

The complete curation pipeline and its stage-specific prompts are provided in the \textit{Code folder} of the Code and Data Supplementary materials, while the \textit{Data folder} provides the metadata, source-video URLs, and timestamps required to reconstruct SemComp-Data.

\begin{table*}[!htbp]
    \centering
    \small
    \renewcommand{\arraystretch}{1.12}
    \begin{tabular}{
        >{\raggedright\arraybackslash}p{0.28\linewidth}
        >{\raggedright\arraybackslash}p{0.65\linewidth}}
    \toprule
    \textbf{Configuration Group} & \textbf{Keyword Strings} \\
    \midrule
    \texttt{news\_keywords} (13)
    & News; Briefing; Coverage; Current Affairs; Deep Dive; Documentary;
    Exclusive; Headline; Interview; Live; Live Update; Press Conference; Report. \\
    \midrule
    \texttt{movie\_keywords} (13)
    & Behind the scenes; Blooper; Casting; Movie Clip; Playback; Preview; Review;
    Spoiler; Teaser; TV Series; Talk; Talks; TalkShow. \\
    \midrule
    \texttt{entertainment\_keywords} (19)
    & ASMR; Celebrity; Challenge; Concert; Gaming; Highlights; Music Video;
    Podcast; Prank; Reaction; Reacts; Travel; Unboxing; Variety Show; Vlog; HBO;
    SportsCenter; NFL; NBA. \\
    \bottomrule
    \end{tabular}
    \caption{Complete title-filtering keyword configuration.}
\label{tab:supp-title-keywords}
\end{table*}

\section{Title-Based Keyword Filtering}\label{supp:title-filtering}
Title-based filtering is used to remove narration-dependent or primarily informational videos whose outcomes cannot be reliably verified from visual evidence alone. 
Such videos often require spoken explanations, contextual knowledge, or other nonvisual information to determine whether the depicted task has been completed successfully, making them unsuitable for visually grounded evaluation. 
To ensure that this filtering step is transparent and reproducible, the implementation loads 45 keyword strings from the filtering configuration. 
Although these keywords are organized into three groups for readability and maintenance, the groups are flattened into a single list before matching.
Each candidate title is checked against every keyword using direct, case-sensitive substring matching, as implemented in the released code.
A video is excluded whenever its title contains at least one listed string with matching capitalization; an exact match to the entire title is not required.
This simple deterministic procedure ensures that the same filtering criterion is applied consistently across all candidate videos. 
The complete filtering configuration is reported in Table~\ref{tab:supp-title-keywords}.

\section{Domain and Category Taxonomy}\label{supp:taxonomy}
SemComp-Data organizes its instances into six broad domains and 21 fine-grained categories.
The domains characterize general application scenarios, whereas the categories distinguish task-specific reference-to-outcome patterns.
Table~\ref{tab:supp-category} lists the complete taxonomy.


\begin{table*}[ht]
    \centering
    \small
    \setlength{\tabcolsep}{5pt}
    \renewcommand{\arraystretch}{1.00}
    \begin{tabular}{
        >{\raggedright\arraybackslash}p{0.17\linewidth}
        >{\raggedright\arraybackslash}p{0.20\linewidth}
        >{\raggedright\arraybackslash}p{0.56\linewidth}}
    
    \toprule
    \textbf{Domain} & \textbf{Category} & \textbf{Definition} \\
    \midrule
    \multirow{3}{=}{Food and Cooking}
    & Dish Making & A dish or food product is completed from ingredients. \\
    & Food Transformation & The same food item undergoes a visible cooking-related state change. \\
    & Food Plating & Food elements are arranged into a complete presentation. \\
    
    \midrule
    \multirow{4}{=}{Beauty and Fashion}
    & Styling & The appearance of the same person is visibly restyled. \\
    & Tool Cleaning & A beauty or fashion tool is cleaned to a reusable condition. \\
    & Try-on & An external appearance reference is applied to a subject. \\
    & Product Reveal & A previously concealed product becomes visible. \\
    
    \midrule
    \multirow{3}{=}{Sports and Fitness}
    & Gear Setup & Sports gear or equipment is configured into a ready-to-use state. \\
    & Body Preparation & The subject's body enters a sports-related prepared state. \\
    & Sports Edits & Source event footage or identity cues are transformed into a completed sports-media edit. \\
    
    \midrule
    \multirow{5}{=}{Crafts and DIY}
    & Assembly & Separate parts are combined into a complete structure or object. \\
    & Restoration & The condition of the same object is restored or refurbished. \\
    & Renovation & A built space is reorganized or remodeled into a new spatial state. \\
    & Blueprint Construction & A physical structure is built according to an abstract design specification. \\
    & Material Shaping & A raw material is physically reshaped into a new form. \\
    
    \midrule
    \multirow{3}{=}{Gardening and Pets}
    & Plant Treatment & A plant is improved through care-related operations. \\
    & Floral Design & Floral or plant elements are trimmed and arranged into an aesthetic design. \\
    & Pet Grooming & A pet's appearance is improved through grooming. \\
    
    \midrule
    \multirow{3}{=}{Arts and Precision}
    & Artwork Creation & An artwork is completed from an unfinished visual basis. \\
    & Digital Creation & A digital creative product is completed from a visual or abstract reference. \\
    & Sculpting & A sculptable material is turned into a finished three-dimensional artwork. \\
    \bottomrule
    
    \end{tabular}
    \caption{Domain--category taxonomy and category definitions used in
    SemComp-Data.}
\label{tab:supp-category}
\end{table*}

\begin{table*}[t]
    \centering
    \small
    \setlength{\tabcolsep}{5pt}
    \renewcommand{\arraystretch}{1.00}
    \begin{tabular}{
        >{\raggedright\arraybackslash}p{0.19\linewidth}
        >{\raggedright\arraybackslash}p{0.37\linewidth}
        >{\raggedright\arraybackslash}p{0.37\linewidth}}
    \toprule
    \textbf{Category} & \textbf{Reference State} & \textbf{Outcome State} \\
    \midrule
    Dish Making
    & Ingredients or an incomplete food preparation are visible.
    & A recognizable completed dish or food product is visible. \\
    
    Food Transformation
    & A food item is visible before a cooking-related state change.
    & The same food item exhibits the intended visible state change. \\
    
    Food Plating
    & Food components are present before final arrangement.
    & The food components form a complete plated presentation. \\
    
    Styling
    & A person is visible before the target styling change.
    & The same person visibly exhibits the completed styling result. \\
    
    Tool Cleaning
    & A beauty or fashion tool is visibly soiled or not ready for reuse.
    & The same tool is visibly clean and reusable. \\
    
    Try-on
    & The subject and an external appearance reference are identifiable.
    & The subject visibly exhibits the referenced garment, accessory, or appearance. \\
    
    Product Reveal
    & The product is concealed, covered, or not yet identifiable.
    & The product is exposed and visually identifiable. \\
    
    Gear Setup
    & Sports gear is unconfigured, disassembled, or not ready for use.
    & The gear is visibly configured in a ready-to-use state. \\
    
    Body Preparation
    & A subject is visible before completing a sports-related preparation.
    & The subject visibly reaches the intended prepared state. \\
    
    Sports Edits
    & Source event footage or identity cues for a sports edit are visible.
    & A completed sports-media edit is visible. \\
    
    Assembly
    & Separate or partially combined components are visible.
    & The components form a complete structure or object. \\
    
    Restoration
    & A worn, damaged, or degraded object is visible.
    & The same object is visibly restored or refurbished. \\
    
    Renovation
    & A built space is visible before reorganization or remodeling.
    & The space exhibits the completed renovated layout or appearance. \\
    
    Blueprint Construction
    & An abstract design, plan, or blueprint is visible.
    & A corresponding completed physical structure is visible. \\
    
    Material Shaping
    & Raw or incompletely shaped material is visible.
    & The material has the intended completed form. \\
    
    Plant Treatment
    & A plant is visible before care or treatment.
    & The plant exhibits a visibly improved state after treatment. \\
    
    Floral Design
    & Unarranged floral or plant elements are visible.
    & The elements form a completed aesthetic arrangement. \\
    
    Pet Grooming
    & A pet is visible before grooming.
    & The same pet exhibits the completed grooming result. \\
    
    Artwork Creation
    & An unfinished artwork or visual basis is visible.
    & A completed artwork is visible. \\
    
    Digital Creation
    & A visual reference or unfinished digital artifact is visible.
    & A completed digital creative product is visible. \\
    
    Sculpting
    & Raw or partially shaped sculptable material is visible.
    & A completed three-dimensional sculpture is visible. \\
    \bottomrule
    \end{tabular}
    \caption{Category-specific reference and outcome state definitions used for
    State Mining.}
\label{tab:supp-states}
\end{table*}

\section{Reference and Outcome State Definitions}\label{supp:state-definitions}
The construction pipeline uses category-specific state definitions to localize representative reference and outcome frames. 
A reference state describes the initial visual evidence required to ground the task, while an outcome state
describes the visually verifiable completed result. Table~\ref{tab:supp-states} summarizes the definitions used for State Mining.

\section{Alignment and Attributes}\label{supp:alignment}

Instruction Structuring first assigns one of four alignment types and then selects instance-specific attributes to preserve from the reference image.
The alignment type identifies the principal form of semantic grounding in the reference; it does not require all attributes associated with that type to remain unchanged.
Table~\ref{tab:supp-alignment-types} gives the operational definitions.
The complete preservation-attribute vocabulary contains 17 options spanning object semantics, visual appearance, human identity, and scene structure.

\begin{table}[!htbp]
    \centering
    \small
    \renewcommand{\arraystretch}{1.00}
    \begin{tabular}{
        >{\raggedright\arraybackslash}p{0.18\linewidth}
        >{\raggedright\arraybackslash}p{0.43\linewidth}
        >{\raggedright\arraybackslash}p{0.31\linewidth}}
    \toprule
    \textbf{Alignment Type}
    & \textbf{Operational Definition}
    & \textbf{Typical Relevant Attributes} \\
    \midrule
    Object Element
    & Grounds the outcome in task-relevant object categories, components,
    ingredients, materials, quantities, or colors appearing in the reference,
    without requiring the exact appearance of a specific object instance.
    & Object category, composition, count, material, and color palette. \\
    \midrule
    
    Identity
    & Requires the depicted person to remain identifiable across the reference
    and generated outcome.
    & Person identity, face appearance, body appearance, and pose. \\
    \midrule
    
    Object Appearance
    & Grounds the outcome in the instance-level visual appearance of a specific
    reference object.
    & Object category, color palette, material, shape structure, surface
    appearance, and graphic details. \\
    \midrule
    
    Scene
    & Grounds the outcome in the organization or context of the reference scene.
    & Scene type, scene layout, spatial relation, background context, and
    viewpoint. \\
    \midrule
    \multicolumn{3}{>{\raggedright\arraybackslash}p{0.94\linewidth}}{
    \textbf{Complete preservation-attribute vocabulary (17):}
    \texttt{object\_category}, \texttt{object\_composition},
    \texttt{object\_count}, \texttt{material}, \texttt{color\_palette},
    \texttt{shape\_structure}, \texttt{surface\_appearance},
    \texttt{graphic\_details}, \texttt{person\_identity},
    \texttt{face\_appearance}, \texttt{body\_appearance}, \texttt{pose},
    \texttt{scene\_type}, \texttt{scene\_layout},
    \texttt{spatial\_relation}, \texttt{background\_context}, and
    \texttt{viewpoint}.} \\
    \bottomrule
    \end{tabular}
    \caption{Operational definitions of the four alignment types and the complete
    preservation-attribute vocabulary. Attribute selection remains
    instance-specific.}
\label{tab:supp-alignment-types}
\end{table}

\section{SemComp-Bench Evaluation Prompts}\label{supp:evaluation-prompts}

\begin{figure*}[t]
    \centering
    \includegraphics[width=1.0\textwidth]{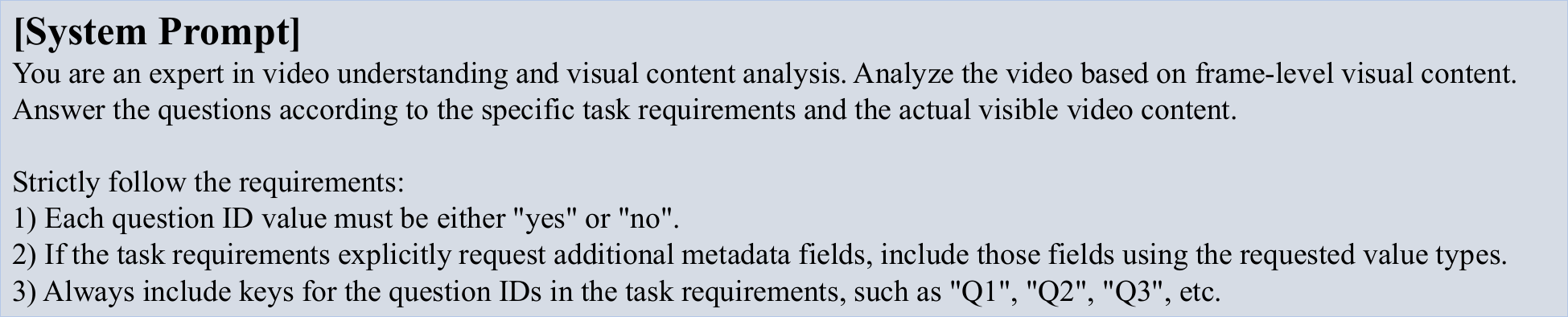}
    \caption{Common system prompt used for SemComp-Bench evaluation.}
    \label{fig:supp-system-prompt}
\end{figure*}

For each generated video, 27 frames are uniformly sampled and presented in temporal order. 
The evaluator answers each question with \emph{yes} or \emph{no} and provides visual evidence for the decision. The two positive OA questions pass on \emph{yes}; the remaining OA questions and all GR questions are failure-oriented and pass on \emph{no}.
SemComp-Bench combines a common system prompt with criterion-specific task prompts. 
The system prompt in Fig.~\ref{fig:supp-system-prompt} defines the evaluator role, restricts each binary decision to \emph{yes} or \emph{no}, and requires all requested question identifiers to appear in the output.

\begin{figure*}[t]
    \centering
    \includegraphics[width=1.0\textwidth]{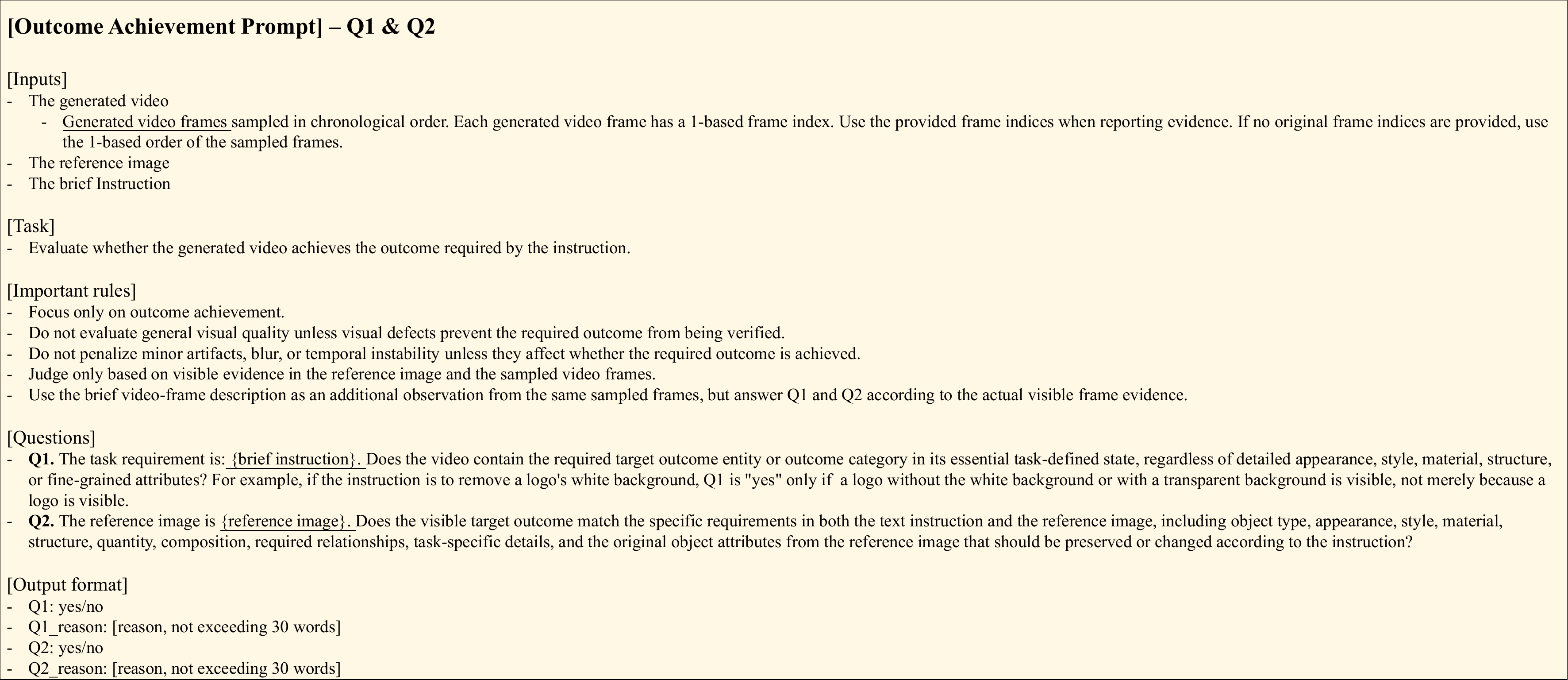}
    \caption{Outcome Achievement prompt for Outcome Realization
    ($Q_{\mathrm{OA}}^1$) and Semantic Grounding ($Q_{\mathrm{OA}}^2$).}
    \label{fig:supp-oa-prompt-q12}
\end{figure*}

\begin{figure*}[t]
    \centering
    \includegraphics[width=1.0\textwidth]{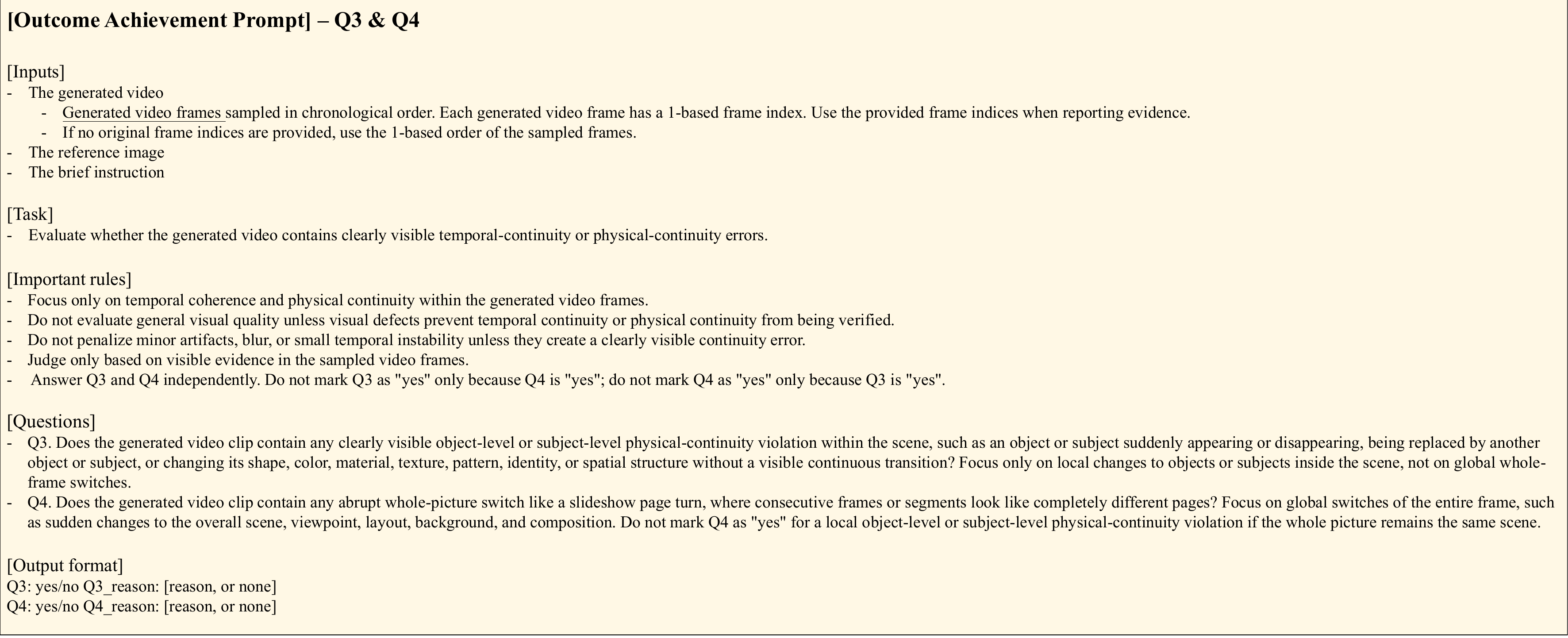}
    \caption{Outcome Achievement prompt for Grounded Entity Consistency
    ($Q_{\mathrm{OA}}^3$) and Global Visual Continuity
    ($Q_{\mathrm{OA}}^4$).}
    \label{fig:supp-oa-prompt-q34}
\end{figure*}

\begin{figure*}[t]
    \centering
    \includegraphics[width=1.0\textwidth]{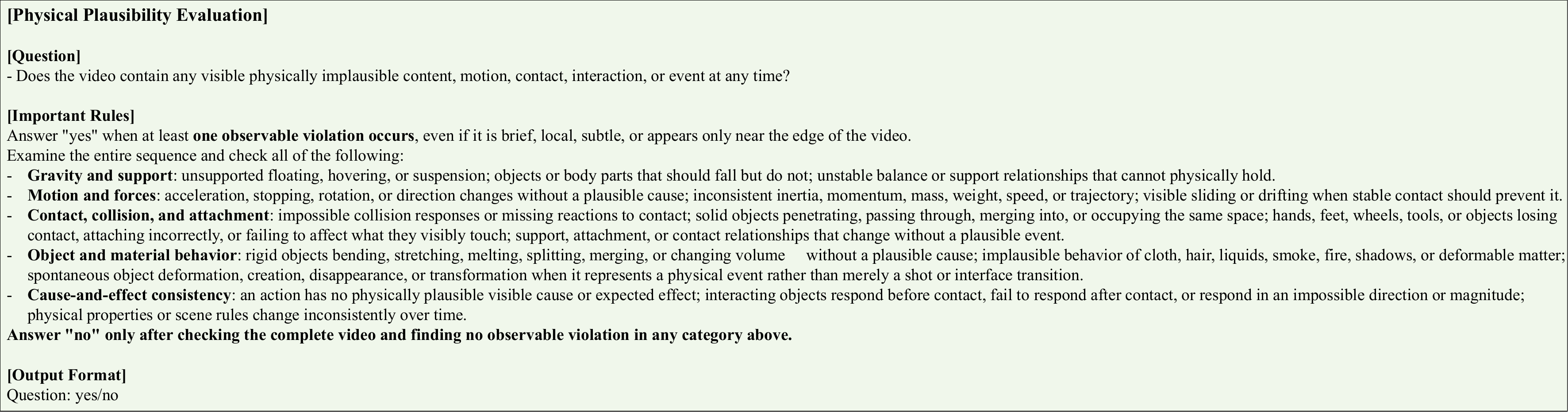}
    \caption{Generation Reliability prompt for Physical Plausibility
    ($Q_{\mathrm{GR}}^1$).}
    \label{fig:supp-gr-prompt-q1}
\end{figure*}

\begin{figure*}[t]
    \centering
    \includegraphics[width=1.0\textwidth]{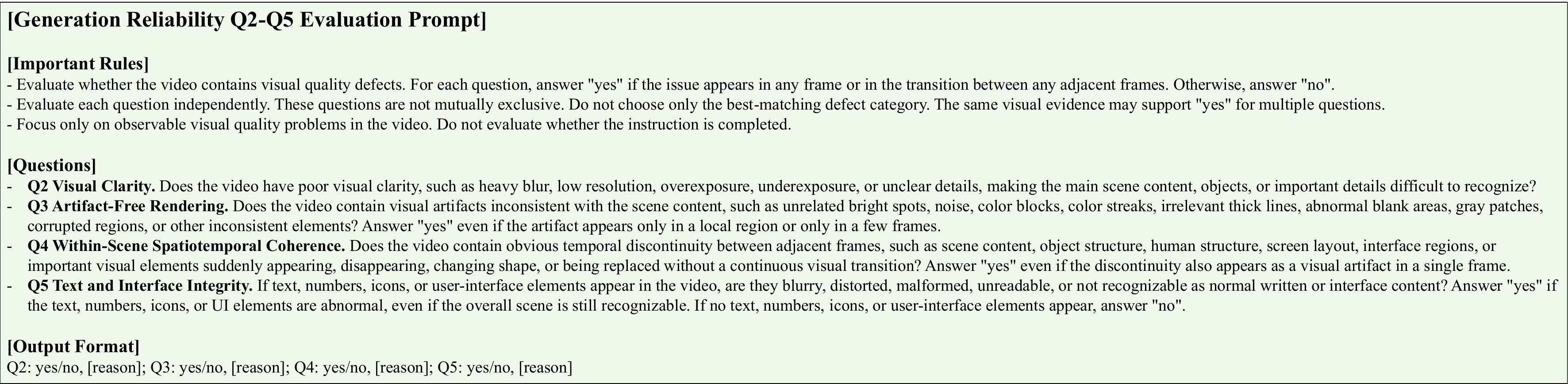}
    \caption{Generation Reliability prompt for Visual Clarity,
    Artifact-Free Rendering, Within-Scene Spatiotemporal Coherence, and Text
    and Interface Integrity ($Q_{\mathrm{GR}}^2$--$Q_{\mathrm{GR}}^5$).}
    \label{fig:supp-gr-prompt-q25}
\end{figure*}

Outcome Achievement is evaluated using two prompt groups. The first prompt, shown in Fig.~\ref{fig:supp-oa-prompt-q12}, evaluates Outcome Realization and Semantic Grounding through $Q_{\mathrm{OA}}^1$ and $Q_{\mathrm{OA}}^2$. 
A brief video-frame description generated by the VLM during State Mining is supplied as auxiliary context; the binary decisions remain grounded in the visible sampled frames.
The second prompt, shown in Fig.~\ref{fig:supp-oa-prompt-q34}, evaluates Grounded Entity Consistency and Global Visual Continuity through $Q_{\mathrm{OA}}^3$ and $Q_{\mathrm{OA}}^4$. 
Grounded Entity Consistency checks for unexplained disappearance, replacement, or task-relevant semantic drift of key entities, whereas Global Visual Continuity checks for abrupt whole-frame changes.
These questions are evaluated from temporally ordered sampled frames, with the reference image and instruction additionally provided when required by the criterion.
Together, the four OA criteria separate task fulfillment from outcome-video validity.
Outcome Realization first establishes whether the instructed completed state is visibly present at a coarse semantic level, while Semantic Grounding checks the task-relevant correspondence between that outcome and the reference--instruction pair.
Grounded Entity Consistency does not require pixel-level appearance constancy; it only flags unexplained changes that prevent key entities from remaining semantically identifiable. 
Global Visual Continuity instead operates at the whole-frame level and detects abrupt switches in the scene, viewpoint, layout, background, or composition. Neither safeguard requires the video to reproduce the intermediate procedure.
Generation Reliability uses a dedicated standalone prompt for Physical Plausibility and a joint prompt for the remaining diagnostic criteria.
Figure~\ref{fig:supp-gr-prompt-q1} shows the standalone prompt for $Q_{\mathrm{GR}}^1$, and Fig.~\ref{fig:supp-gr-prompt-q25} shows the joint prompt for $Q_{\mathrm{GR}}^2$--$Q_{\mathrm{GR}}^5$, covering Visual Clarity, Artifact-Free Rendering, Within-Scene Spatiotemporal Coherence, and Text and
Interface Integrity. 
All five GR questions are failure-oriented, so a \emph{no} response denotes a pass.
GR is assessed independently of task completion and semantic grounding. 
Physical Plausibility examines visible motion, contact, interaction, and material behavior across the full temporally ordered sampled sequence. 
The remaining questions distinguish recognizability problems, scene-inconsistent rendering artifacts, local instability within otherwise continuous scenes, and malformed text or interface elements. 
Because these failure types are not mutually exclusive, the evaluator answers each question independently and may cite the same visual evidence for more than one criterion.
Criterion-specific answers and concise evidence keep the reported pass rates interpretable and consistent with the main-paper score definitions.

\FloatBarrier

\onecolumn

\section{Dataset Statistics and Instances} \label{supp:statistics}

The curation pipeline yields 1,273 SemComp-Data instances. Figure~\ref{fig:supp-data-instances} presents 22 additional instances spanning diverse task categories. 
Each example shows the reference image, three temporally ordered frames from the paired outcome-centric clip, and the brief instruction.

\begin{center}
    \centering
    \includegraphics[width=1.0\textwidth]{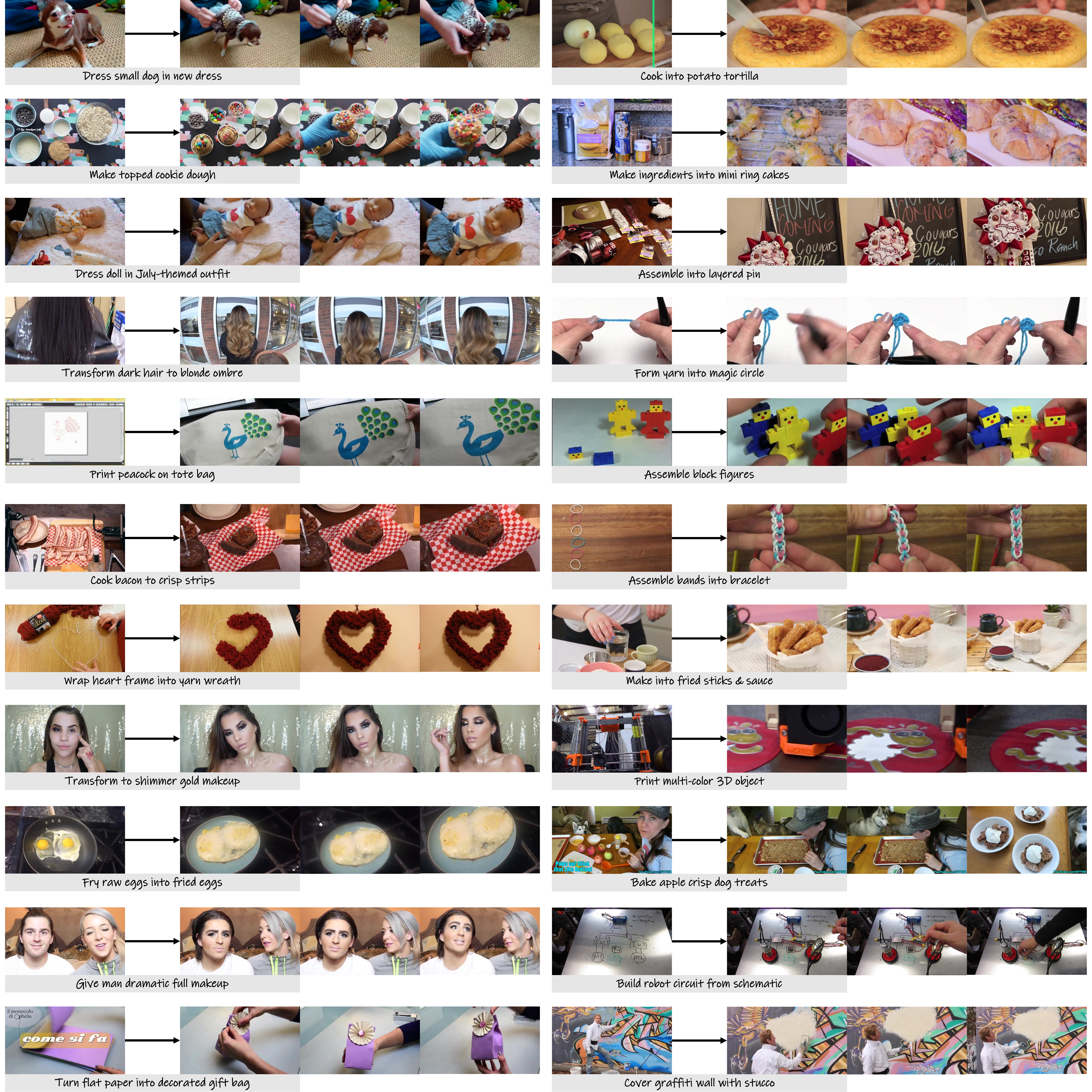}
    \captionof{figure}{Additional SemComp-Data instances. Each example shows a reference
    image, three temporally ordered frames sampled from the paired
    outcome-centric video clip, and the corresponding brief instruction.}
    \label{fig:supp-data-instances}
\end{center}

\noindent
\begin{minipage}[t]{0.59\textwidth}
\vspace{0pt}
The arrows denote the intended semantic transitions to the completed outcomes. 
This notation does not imply that a generated video must reproduce the intermediate procedures.
The splitting module first partitions each full-context video into shots and merges temporally adjacent, visually consistent shots into events.
The event containing the verified outcome timestamp is then selected, after which duration-constrained extraction produces the outcome-centric clip.
Events within the nominal target range of $3$--$4\,\mathrm{s}$ are retained in full; otherwise, an outcome-centered window is extracted from the full-context video with a target duration of $4\,\mathrm{s}$ for an overlong event or $3\,\mathrm{s}$ for an underlength event, with boundary-aware temporal shifting when necessary. 
This procedure results in a mean clip duration of approximately $4.03\,\mathrm{s}$.
\end{minipage}\hfill
\begin{minipage}[t]{0.37\textwidth}
\vspace{0pt}
\centering
\small
\begin{tabular}{lr}
    \toprule
    \textbf{Statistic} & \textbf{Value} \\
    \midrule
    Number of instances & 1,273 \\
    Minimum clip duration & $3.020\,\mathrm{s}$ \\
    Median clip duration & $4.067\,\mathrm{s}$ \\
    Mean clip duration & $4.033\,\mathrm{s}$ \\
    Maximum clip duration & $4.125\,\mathrm{s}$ \\
    SemComp-Core instances & 60 \\
    Sampled evaluation frames & 27 \\
    \bottomrule
\end{tabular}
\captionof{table}{Additional dataset and evaluation statistics.}
\label{tab:supp-duration}
\end{minipage}

\FloatBarrier

\section{Qualitative Results}\label{supp:qualitative}

\begin{figure}[!htbp]
    \centering
    \includegraphics[width=1.0\textwidth]{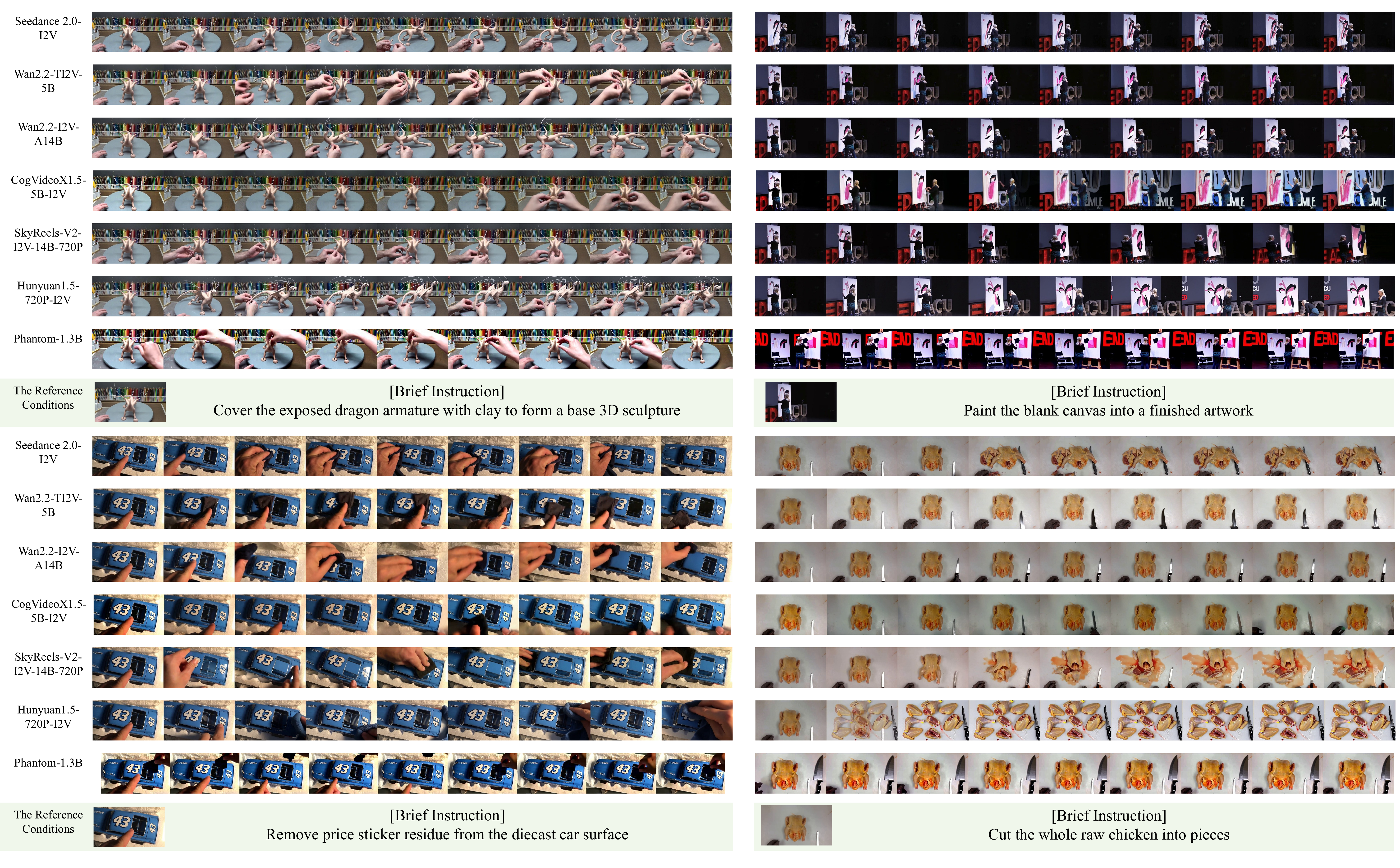}
    \caption{Additional generated-video comparisons on four SemComp-Core
    instances. Outputs from all seven I2V models in each block were generated using
    the shared reference condition and detailed instruction; the paired brief instruction
    is displayed only for compact presentation. Temporally ordered output frames
    illustrate differences in both outcome achievement and generation reliability.}
    \label{fig:supp-model-output-2}
\end{figure}

\begingroup
\setlength{\emergencystretch}{1em}
Figure~\ref{fig:supp-model-output-2} provides four additional SemComp-Core comparisons spanning Sculpting, Artwork Creation, Restoration, and Food Transformation.
Within each case, all seven models receive the same reference condition and detailed instruction; the paired brief instruction is displayed in the figure only for compact presentation. Their outputs are shown as temporally ordered frame sequences.
These matched comparisons expose complementary failure modes that are not fully captured by a terminal frame alone. 
Some sequences preserve the reference scene but make limited progress toward the specified outcome, whereas others depict a partial state change while introducing entity deformation, material drift, or abrupt visual transitions.
The examples therefore illustrate why Outcome Achievement and Generation Reliability provide complementary views of semantic task completion.
\par
\endgroup

\FloatBarrier

\section{Result Variability}\label{supp:result-std}

Each generated video is evaluated by three independent VLM calls made at different times. Tables~\ref{tab:supp-oa-std}--\ref{tab:supp-conditioning-std} report the sample standard deviation of the three corresponding run-level scores. 
Specifically, for run-level score $x_r$, we compute $\sigma(x)=\sqrt{\frac{\sum_{r=1}^{3}(x_r-\bar{x})^2}{3-1}}$. All entries are measured in percentage points (pp).
The OA Score standard deviations use the conjunctive definition in which a sample must pass all four OA criteria, whereas the GR Score standard deviations are computed from the run-level mean of the five GR
criterion pass rates. 
Thus, the aggregate-score columns follow exactly the definitions used for the results reported in the main paper.

\begin{table}[!htbp]
    \centering
    \footnotesize
    \setlength{\tabcolsep}{3pt}
    \renewcommand{\arraystretch}{1.12}
    \begin{tabular}{
        >{\raggedright\arraybackslash}p{0.34\textwidth}
        *{5}{>{\centering\arraybackslash}p{0.105\textwidth}}}
    \toprule
    \textbf{Model}
    & $\sigma(A_{\mathrm{or}})$
    & $\sigma(A_{\mathrm{sg}})$
    & $\sigma(A_{\mathrm{gec}})$
    & $\sigma(A_{\mathrm{gvc}})$
    & $\sigma(\text{OA Score})$ \\
    \midrule
    Seedance 2.0                 & 0.96 & 0.96 & 2.55 & 7.52 & 4.41 \\
    Wan2.2-TI2V-5B              & 4.19 & 4.41 & 3.85 & 0.96 & 3.33 \\
    Wan2.2-I2V-A14B             & 1.67 & 2.55 & 5.09 & 3.47 & 2.89 \\
    CogVideoX1.5-5B-I2V         & 3.33 & 4.19 & 2.55 & 0.96 & 0.96 \\
    SkyReels-V2-I2V-14B-720P    & 1.67 & 3.47 & 3.47 & 2.55 & 0.96 \\
    HY$^\dagger$-1.5-720P-I2V  & 2.55 & 2.55 & 2.89 & 1.92 & 3.47 \\
    Phantom-1.3B                 & 0.96 & 1.92 & 1.92 & 0.96 & 0.96 \\
    \bottomrule
    \end{tabular}
    \caption{Sample standard deviations for SemComp-Core Outcome
    Achievement under detailed instructions (pp; three runs). $^\dagger$HY
    denotes HunyuanVideo.}
    \label{tab:supp-oa-std}
\end{table}

\begin{table}[!htbp]
    \centering
    \footnotesize
    \setlength{\tabcolsep}{3pt}
    \renewcommand{\arraystretch}{1.12}
    \begin{tabular}{
        >{\raggedright\arraybackslash}p{0.32\textwidth}
        *{5}{>{\centering\arraybackslash}p{0.085\textwidth}}
        >{\centering\arraybackslash}p{0.10\textwidth}}
    \toprule
    \textbf{Model}
    & $\sigma(G_{\mathrm{pp}})$
    & $\sigma(G_{\mathrm{vc}})$
    & $\sigma(G_{\mathrm{afr}})$
    & $\sigma(G_{\mathrm{wsc}})$
    & $\sigma(G_{\mathrm{ti}})$
    & $\sigma(\text{GR Score})$ \\
    \midrule
    Seedance 2.0                 & 3.33 & 0.96 & 0.96 & 1.92 & 0.96 & 1.58 \\
    Wan2.2-TI2V-5B              & 0.96 & 2.55 & 6.74 & 12.95 & 0.96 & 3.95 \\
    Wan2.2-I2V-A14B             & 5.85 & 0.96 & 1.67 & 1.92 & 4.19 & 1.53 \\
    CogVideoX1.5-5B-I2V         & 2.55 & 13.47 & 6.31 & 8.82 & 8.55 & 7.12 \\
    SkyReels-V2-I2V-14B-720P    & 0.96 & 3.47 & 6.74 & 11.10 & 3.85 & 4.53 \\
    HY$^\dagger$-1.5-720P-I2V  & 8.82 & 1.92 & 0.00 & 6.94 & 0.96 & 1.35 \\
    Phantom-1.3B                 & 5.09 & 0.96 & 8.22 & 15.84 & 6.94 & 7.20 \\
    \bottomrule
    \end{tabular}
    \caption{Sample standard deviations across three evaluation runs for the
    SemComp-Core Generation Reliability results under detailed instructions.
    All values are in percentage points. $^\dagger$HY denotes HunyuanVideo.}
    \label{tab:supp-gr-std}
\end{table}

\begin{table}[!htbp]
    \centering
    \footnotesize
    \setlength{\tabcolsep}{2.5pt}
    \renewcommand{\arraystretch}{1.10}
    \begin{tabular}{
        >{\raggedright\arraybackslash}p{0.21\textwidth}
        >{\centering\arraybackslash}p{0.075\textwidth}
        >{\centering\arraybackslash}p{0.10\textwidth}
        *{4}{>{\centering\arraybackslash}p{0.085\textwidth}}
        >{\centering\arraybackslash}p{0.105\textwidth}}
    \toprule
    \textbf{Model} & \textbf{Modality} & \textbf{Instruction}
    & $\sigma(A_{\mathrm{or}})$
    & $\sigma(A_{\mathrm{sg}})$
    & $\sigma(A_{\mathrm{gec}})$
    & $\sigma(A_{\mathrm{gvc}})$
    & $\sigma(\text{OA Score})$ \\
    \midrule
    \multirow{3}{=}{Wan2.2-A14B}
    & I2V & Detailed & 1.67 & 2.55 & 5.09 & 3.47 & 2.89 \\
    & T2V & Detailed & 2.55 & 4.19 & 1.67 & 4.41 & 2.55 \\
    & T2V & Brief    & 0.96 & 0.96 & 2.55 & 7.26 & 0.96 \\
    \midrule
    \multirow{3}{=}{CogVideoX1.5-5B}
    & I2V & Detailed & 3.33 & 4.19 & 2.55 & 0.96 & 0.96 \\
    & T2V & Detailed & 3.33 & 0.96 & 2.55 & 0.96 & 1.67 \\
    & T2V & Brief    & 2.89 & 1.67 & 0.96 & 3.47 & 0.96 \\
    \midrule
    \multirow{3}{=}{HY$^\dagger$-1.5-720P}
    & I2V & Detailed & 2.55 & 2.55 & 2.89 & 1.92 & 3.47 \\
    & T2V & Detailed & 0.00 & 2.55 & 6.74 & 5.00 & 1.92 \\
    & T2V & Brief    & 1.67 & 3.33 & 4.19 & 5.36 & 0.00 \\
    \bottomrule
    \end{tabular}
    \caption{Sample standard deviations across three evaluation runs for the
    Outcome Achievement conditioning study. All values are in percentage
    points. $^\dagger$HY denotes HunyuanVideo.}
    \label{tab:supp-conditioning-std}
\end{table}

\end{document}